# AI and Machine Learning Driven Indoor Localization and Navigation with Mobile Embedded Systems

Sudeep Pasricha
*Department of Electrical and Computer Engineering*
Colorado State University, Fort Collins, CO, USA
sudeep@colostate.edu

*Abstract* – Indoor navigation is a foundational technology to assist the tracking and localization of humans, autonomous vehicles, drones, and robots in indoor spaces. Due to the lack of penetration of GPS signals in buildings, subterranean locales, and dense urban environments, indoor navigation solutions typically make use of ubiquitous wireless signals (e.g., WiFi) and sensors in mobile embedded systems to perform tracking and localization. This article provides an overview of the many challenges facing state-of-the-art indoor navigation solutions, and then describes how AI algorithms deployed on mobile embedded systems can overcome these challenges.

## I. Indoor Navigation: Overview

The indoor positioning and navigation market was valued at $25.68 billion USD in 2024 and is expected to reach $174.02 billion USD by 2030 [1]. This growth is being driven by the need for accurate and real-time localization across complex environments for applications such as seamless smartphone-based human navigation (much like how we navigate in outdoor spaces), as well as localization and tracking for robots, drones, autonomous vehicles, and augmented reality (AR) use-cases [2]. While navigation in outdoor spaces today relies on various satellite based Global Navigation Satellite System (GNSS) solutions such as the GPS, no standardized solution exists for GNSS-deficient locales such as buildings, dense urban areas, and underground mines. Thus, researchers across industry and academia have been exploring a variety of indoor localization strategies that rely on different wireless signals (e.g., WiFi, BLE) and sensors.

In general, all indoor localization techniques are based on one of three categories of approaches based on: 1) static propagation models, 2) triangulation/trilateration, or 3) fingerprinting. Early indoor localization solutions used static propagation models that captured the relationship between distance and wireless signal strengths to localize a person carrying a wireless receiver device with respect to a wireless transmitter deployed at a fixed location [3]. Triangulation/trilateration-based methods improved on these approaches by using geometric properties such as the distance between multiple wireless transceivers (aka access points or APs) and a user-carried smartphone, or the angles at which signals from two or more APs are received by the smartphone [4]. But both of these approaches suffer from degraded accuracies when scaled up across large and complex indoor environments, where multipath and shadowing effects due to walls and other obstacles create large errors. As fingerprinting is much more resilient to these effects, most recent indoor localization and navigation solutions, as well as competitions from NIST and Microsoft [5], have relied on it.

Fingerprinting is based on the idea that each indoor location exhibits a unique signature of AP characteristics, e.g., received signal strength indicator (RSSI) values [6]. Thus, in a building with 15 WiFi APs, each location can be characterized with a unique vector of 15 WiFi RSSI values (fingerprint) that can be easily captured by a commodity smartphone. The fingerprint at a location is a unique identifier for it. Fingerprinting is usually carried out in two phases. In the first (offline) phase, RSSI values are collected along a finite number of indoor locales. The resulting database of values is then often used to train models (e.g., machine learning-based) for location estimation. In the second (online) phase, the models are deployed on user-carried devices (e.g., smartphones) and used to predict user location. Interactive navigation can be supported based on real-time readings of AP RSSI values on the smartphone. State-of-the-art solutions typically make use of WiFi RSSI, due to the ubiquitous deployment of WiFi APs in most indoor locales [2].

## II. Challenges with Indoor Navigation

Indoor navigation requires real-time localization with consistent accuracies of a few meters or less, which is challenging due to many types of obstacles such as walls, furniture, electronic equipment, and stationary or moving humans, all of which give rise to multipath effects in wireless signals due to signal reflection, attenuation, and noise interference. The variety of materials used within indoor locales (metals, concrete, wood, plastics) further increases wireless signal propagation uncertainty. Compounding this challenge is the fact that different smartphones can have diverse hardware and software stacks and two different smartphones at the same location can end up capturing dissimilar RSSI values. State-of-the-art WiFi fingerprinting solutions thus need to prioritize achieving consistent *high localization accuracy in the presence of device heterogeneity and environmental uncertainties*. Indoor navigation must also support *real-time location prediction* for seamless navigation, which is not straightforward with resource-limited smartphones that have performance constraints. Mobile devices are also battery-based with a limited energy supply, thus solutions must ensure *energy-efficient deployment*. Another challenge is the ability to *scale up and predict locations efficiently* across large indoor environments with high user counts. The deployed indoor navigation solution must also maintain *long-term stability of predictions* over lengthy durations of time, e.g., months to years after deployment. Lastly, the solutions must be able to overcome security challenges due to malicious attackers that aim to compromise the deployed solution for nefarious goals.

## III. AI-Driven Solutions for Indoor Navigation

Fortunately, over the past decade, AI and machine learning have played a pivotal role towards overcoming the many challenges with WiFi fingerprinting based indoor navigation, as discussed next.

One of the earliest frameworks that explored the use of machine learning (ML) for WiFi fingerprinting based indoor navigation with mobile embedded devices was *LearnLoc* [7] in 2015. This framework explored multiple ML algorithms including K Nearest Neighbor (KNN) and Deep Neural Networks (DNN) deployed on Android smartphones and found that the DNN approach provided the highest accuracy that varied between 1 to 3 meters on average. *LearnLoc* was also the first framework to optimize energy-efficiency of a smartphone-based indoor navigation solution during deployment. The *CNN-LOC* framework [8] improved upon this approach with a unique idea: it advocated for converting captured WiFi RSSI fingerprints into images that were then used to train a lightweight convolutional neural network (CNN) model deployed on smartphones. This approach outperformed DNN based solutions.

The use of increasingly powerful ML models for indoor navigation deployed on smartphones is a challenge as these devices possess limited computational, memory, and battery resources. In [9], the *QuickLoc* framework was proposed to reduce overheads of CNN execution for indoor navigation with smartphones. The framework improved upon *CNN-LOC* by adding support for conditional computing within CNNs. Specifically, support was added for early exits during a forward inference pass with the CNN, where a prediction can be obtained by traversing fewer than the total number of layers in the CNN. An autotuning approach was designed to determine, based on a specific input, which of the early exits could provide accurate predictions with high confidence. Experimental analysis showed that *QuickLoc* could reduce prediction latency by up to 42%, reduce energy consumption by up to 45%, and increase battery lifetime by up to 28%, without notable reduction in prediction accuracy. Another approach to reduce the overhead of ML models for indoor navigation was proposed in the *CHISEL* framework [10]. This model compression framework integrated approaches for simultaneous quantization (reduction of ML model parameter bitwidth) and sparsification (increase in number of zero parameters in ML layers) to trade-off location accuracy with its latency, energy, and memory footprint. One of the more promising generated configurations, when deployed on a smartphone, reduced memory footprint by 81.52%, and also reduced latency by 30.93%, at the cost of sacrificing 4.89% in localization accuracy.

In the presence of significant environmental noise or device heterogeneity, the accuracy of even the best ML-based indoor navigation frameworks can degrade significantly. For instance, [11] showed experimental evidence of an average accuracy drop of 6× across different smartphones running the same indoor navigation framework. The same work proposed the *SHERPA* framework that made use of statistical techniques (Pearson's Cross-Correlation, Z-score) to correlate fingerprints in real-time in the online phase with fingerprints from the offline phase, to make a preliminary location prediction. This prediction is eventually passed through a user motion-based filtering mechanism to achieve device heterogeneity-resilient location prediction. Subsequent efforts improved upon the performance of this framework by using more powerful statistical techniques (Spearman's Correlation Coefficient, Zero Normalized Cross-Correlation) [12] and integrating a Hidden Markov Model for improved tracking [13]. The *ANVIL* framework [14] adapted a multi-head attention based DNN to learn to augment fingerprints to mitigate the effects of device heterogeneity, achieving 35% better accuracy compared to previous frameworks across five different indoor environments and considering six different smartphones. The *VITAL* framework improved on this effort by adapting vision transformers for the same problem [15]. The impact of stacked autoencoders for fingerprint augmentation to improve heterogeneity resilience was explored in the *SANGRIA* framework [16]. Most recently, the *STELLAR* framework [17] jointly addressed device heterogeneity and environmental noise challenges by combining multi-head attention networks with extreme gradient boosting.

A practical indoor navigation framework must also be able to scale up to large environments and demonstrate stable performance over time. The *FedHIL* framework [18] proposed a unique approach that involved a federated distributed learning approach for reducing data (fingerprint) collection overheads, preserving user privacy, and speeding up deployment via distributed and coordinated learning. A novel model aggregation strategy was proposed to improve resilience to device heterogeneity and environmental noise. To improve long-term prediction stability and prevent degradation of localization accuracy over lengthy periods of time, the *STONE* framework [19] used a few-shot learning approach based on Siamese neural networks. The framework also adaptively applied augmentations to fingerprints over time to compensate for changes. The proposed approach delivered up to 40% better localization accuracy over a 15-month period, without requiring any re-training or model updating after the initial deployment.

Securing indoor navigation deployments is crucial to ensure reliable performance in real-world contexts. Malicious actors can disrupt such systems by attacking APs, via jamming, interference, or spoofing attacks. The first studies on this theme were conducted in [20] where it was shown that the worst-case localization error for DNN and CNN based indoor navigation frameworks can be increased by up to 20× by targeting just one AP. A noise-aware fingerprint extrapolation approach was proposed to reduce the impact of such attacks for both DNN and CNN based frameworks. The *CALLOC* framework [21] addressed the challenge of adversarial attacks on deep learning, using a curriculum learning technique to enhance resilience to both attacks and environmental noise. Most recently, the SENTINEL framework [22] made use of capsule neural networks to outperform CALLOC in achieving adversarial attack resilience across diverse attack scenarios.

## IV. CONCLUSIONS

In this paper, we presented a brief overview of some of the key challenges for indoor navigation with mobile embedded systems. Indoor navigation is poised to revolutionize mobility and tracking of humans, autonomous vehicles, robots, and drones across GNSS deficient environments. But there remain a multitude of challenges to overcome to achieve high accuracy localization, real-time and energy-efficient deployment, resilience to device and environmental uncertainty, scalability, long-term stability, and security. We discussed several AI-driven strategies to overcome these challenges. The future is bright for innovation and research in this exciting field that will inevitably transform society for the better.